\documentclass{article}

\usepackage{microtype}
\usepackage{graphicx}
\usepackage{subcaption}
\usepackage{booktabs} 

\usepackage{hyperref}
\usepackage[accepted]{icml2026}

\usepackage{amsmath}
\usepackage{amssymb}
\usepackage{mathtools}
\usepackage{amsthm}

\usepackage{multirow}
\usepackage[table]{xcolor}
\usepackage[ruled,vlined,noend]{algorithm2e}
\usepackage{xcolor}
\usepackage{enumitem}

\usepackage{url}
\usepackage{algorithm}
\usepackage{algorithmic}

\usepackage{graphicx}
\usepackage{booktabs}
\usepackage{amsfonts}
\usepackage{array}
\usepackage{tabularx}
\usepackage{makecell}

\definecolor{lighturlblue}{HTML}{4A90E2}
\hypersetup{
colorlinks=true,
urlcolor=lighturlblue
}
\definecolor{bgblue}{rgb}{0.21,0.49,0.74}
\definecolor{darkred}{RGB}{139,0,0}
\definecolor{darkgreen}{RGB}{0,139,0}

\usepackage[capitalize,noabbrev]{cleveref}

\theoremstyle{plain}

\theoremstyle{definition}

\theoremstyle{remark}

\usepackage[textsize=tiny]{todonotes}

\icmltitlerunning{Memory as Plans: World-Action Modeling with Memory-Grounded Planning}

\begin{document}

  \twocolumn[{%
    \icmltitle{Memory as Plans: World-Action Modeling with Memory-Grounded Planning}

    \icmlsetsymbol{equal}{*}
    \vspace{-2mm}
    \begin{icmlauthorlist}
      \icmlauthor{Sizhe Zhao}{hit}
      \icmlauthor{Haozhe Xie}{ntu}
      \icmlauthor{Weiyu Zhao}{hit}
      \icmlauthor{Chenchu Zhang}{hit}
      \icmlauthor{Huan Wang}{sdu} \\
      \icmlauthor{Chenyang Wang}{hit} 
      \icmlauthor{Qinglin Liu}{hit}
      \icmlauthor{Shengping Zhang$^\dagger$}{hit,hitqd}
    \end{icmlauthorlist}

    \icmlaffiliation{hit}
      {Harbin Institute of Technology, China}
    \icmlaffiliation{ntu}
      {Nanyang Technological University, Singapore}
    \icmlaffiliation{sdu}
      {Shandong University, China}
    \icmlaffiliation{hitqd}
      {Harbin Institute of Technology (Weihai) Qingdao Research Institute, China}

    \icmlcorrespondingauthor
      {Shengping Zhang}
      {s.zhang@hit.edu.cn}

    \vskip 0.09in
    
    \begin{center}
    Project Page: \href{https://sizhezhao.github.io/projects/MaP-WAM/}{%
      \textcolor{lighturlblue}{\nolinkurl{{MaP-WAM}}}%
    }
   \end{center}

    \vskip 0.09in

    \begin{minipage}{\textwidth}
      \centering
      \includegraphics[width=\linewidth]{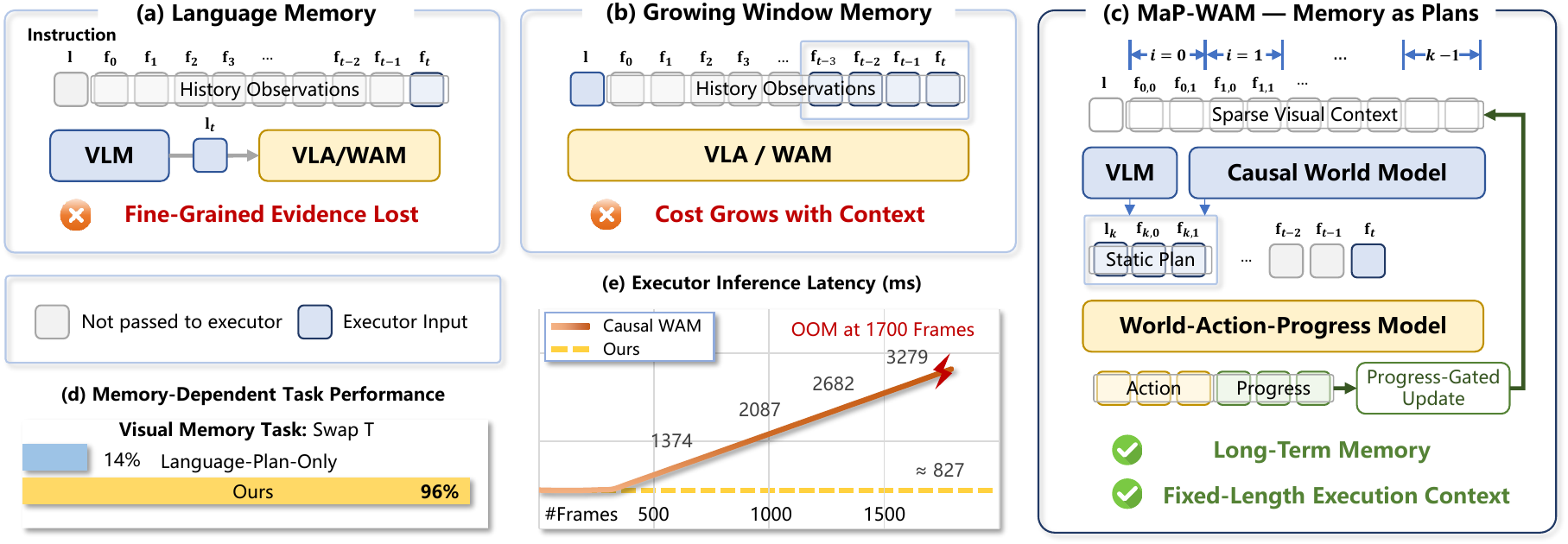}
      \captionof{figure}{\textbf{(a) Language Memory} compactly summarizes past interactions but may discard fine-grained visual evidence, as illustrated in \textbf{(d)}, impairing performance on tasks that require precise visual memory.
  \textbf{(b) Growing Window Memory} retains a window of recent observations, but extending the window to cover longer histories increases executor latency and GPU memory consumption, as shown in \textbf{(e)}, resulting in a trade-off between history coverage and execution efficiency.
  \textbf{(c) MaP-WAM} constructs long-term sparse visual context by retaining a few frames from each completed segment.
  A vision-language model and a causal world model then convert this context into a language-visual plan.
  Conditioned on this static plan, the World-Action-Progress model jointly
  predicts action chunks and progress, enabling adaptive segment transitions and
  memory updates from real observations while keeping the executor context length fixed.}
  \vspace{2mm}
      \label{fig:teaser}
    \end{minipage}

    \vskip 0.1in
}]

\printAffiliationsAndNotice{}

\begin{abstract}
Mainstream robotic policies often adopt a Markovian formulation, but many complex real-world manipulation tasks are inherently non-Markovian, requiring long-horizon memory beyond the current observation.
Existing memory mechanisms often rely on language summaries, growing visual windows, or their combinations, and may therefore lose fine-grained visual evidence or face a trade-off between history coverage and execution efficiency.
We introduce \textbf{MaP-WAM}, a Memory-as-Plans framework that decomposes memory-dependent world-action modeling into memory-grounded planning and plan-conditioned execution, and uses long-term multimodal episodic context as planning-time evidence rather than repeatedly conditioning the executor on the full history.
MaP-WAM represents memory as completed segment records containing language instructions and sparse visual context, and converts this episodic memory into compact plans comprising the next segment-level language plan and corresponding visual guidance.
A World-Action-Progress (WAP) model executes each plan over an unknown duration by jointly predicting action chunks and corresponding execution progress at inference time, calibrating predicted progress through plan-observation alignment for adaptive segment transitions and closed-loop context updates.
MaP-WAM keeps the executor context length fixed, while structured attention further enables key-value caching in both planning and execution.
MaP-WAM achieves state-of-the-art performance on RMBench with an 83.3\% success rate and attains 78.0\% success on real-robot tasks, while maintaining approximately constant executor inference latency as task history grows.
\end{abstract}
\section{Introduction}

Recent advances in vision-language-action (VLA) models~\cite{kim2024openvla, black2024pi_0, black2025pi, wang2025vla} and world-action models (WAMs)~\cite{NEURIPS2023_1d5b9233, hu2025video, kim2025cosmospolicy, yuan2026fastwam} have improved robotic manipulation.
Yet many formulate action prediction under a Markovian assumption, treating the current observation or a fixed short history as sufficient.
This approximation is inadequate for memory-dependent, partially observable tasks, where information required for a future decision may no longer be visible~\cite{shi2026memoryvla, chen2026rmbench, torne2026mem}.
Reliable robotic policies therefore require long-horizon memory beyond the current observation.

Existing memory mechanisms for embodied control often rely on language summaries or growing visual windows (Fig.~\ref{fig:teaser}(a) and (b))~\cite{sridhar2026scaling, chen2026rmbench, torne2026mem, li2026causal, ye2026world, team2026motubrain}.
The former provides compact semantic abstractions but may omit fine-grained visual and spatial evidence.
The latter preserves richer perceptual evidence. Causal WAMs, such as LingBot-VA~\cite{li2026causal}, offer a natural mechanism for retaining long-horizon visual histories by modeling visual dynamics and actions over a growing prefix of episodic observations.
However, conditioning action generation on this frame-wise history incurs increasing computational and GPU-memory costs as the context grows, creating a trade-off between inference efficiency and access to long-horizon history.

We argue that in memory-dependent tasks, long-horizon visual history is not necessarily required as a direct input to the execution model at every control step.
This history is primarily needed to determine the next segment-level plan and the desired visual evolution, while execution can operate by following a memory-grounded plan.
This motivates \textbf{MaP-WAM} (Fig.~\ref{fig:teaser}(c)), which maintains long-term memory as multimodal episodic context and uses it as planning-time evidence to generate compact plans. 
By decoupling memory-grounded planning from plan-conditioned execution, MaP-WAM keeps the executor context length fixed and reduces execution-time latency while preserving fine-grained grounding in long-horizon visual memory. 

Concretely, MaP-WAM maintains multimodal episodic context composed of the task instruction, completed segment instructions, and sparse visual context.
At each planning stage, a language planner predicts the next segment-level language plan, and a causal world model generates corresponding visual guidance conditioned on this plan and the sparse visual context.
Together, the language plan and visual guidance form a memory-grounded plan that couples task semantics with anticipated visual evolution.
Executing this plan, however, poses a central challenge: the required execution duration is initially unknown, depending on task requirements and stochastic execution dynamics.
We therefore introduce a World-Action-Progress (WAP) model that jointly models future visual dynamics, action chunks, and execution progress using a Mixture-of-Transformers (MoT) architecture~\cite{LiangYL0DZGLYZL25}.
The predicted progress enables MaP-WAM to execute each plan for a variable duration and replan upon segment completion.
Progress modeling also equips the executor with an explicit temporal coordinate for distinguishing visually similar observations that correspond to different semantic stages.
Furthermore, the visual plan enables plan-observation alignment, which calibrates predicted progress by matching the current observation to the planned visual trajectory, thereby mitigating cumulative drift of progress prediction over long executions.

The contributions are summarized as follows:

\begin{itemize}
  \item \textbf{Memory-as-Plans Framework}: We propose MaP-WAM, which converts long-horizon episodic evidence into memory-grounded plans, enabling fixed-context execution while retaining visual grounding.

  \item \textbf{Memory-Grounded Planning}: We introduce a causal world model that translates long-term visual context and the predicted segment-level language plan into a visual plan as fine-grained execution guidance.

  \item \textbf{Progress-Aware Execution}: We introduce WAP, which jointly models visual dynamics, action chunks, and execution progress, and combines MoT-based progress prediction with plan-observation alignment to enable variable-duration execution and adaptive segment transitions.

  \item MaP-WAM achieves 83.3\% and 78.0\% success rates on RMBench and real-robot tasks, respectively, while maintaining approximately constant executor inference latency as task history grows.
\end{itemize}

\section{Related Work}

\subsection{Generalist Robotic Policies}

\textbf{Vision-Language-Action Policies.}
Vision-language-action (VLA) policies~\cite{zitkovich2023rt, team2024octo, kim2024openvla, DBLP:journals/corr/abs-2506-01844, black2024pi_0, black2025pi, wang2025vla} leverage semantic priors from pretrained vision-language foundation models~\cite{karamcheti2024prismatic, beyer2024paligemma, bai2025qwen3} and scale policy learning with large-scale datasets~\cite{khazatsky2024droid, o2024open, bu2025agibot}, improving instruction following and task generalization.
However, most VLA policies remain conditioned on the current observation or a fixed short window. Recent designs such as DynamicVLA~\cite{xie2026dynamicvla} further optimize this reactive regime for low-latency control by overlapping inference with execution.
Such formulations are effective for reactive manipulation but struggle with memory-dependent tasks.

\noindent\textbf{World-Action Models.}
World-action models (WAMs) enhance action generation through world modeling~\cite{NEURIPS2023_1d5b9233, hu2025video, kim2025cosmospolicy, yuan2026fastwam, li2026causal, DBLP:journals/corr/abs-2603-10448, ye2026world, team2026motubrain}.
By predicting future latent states, future observations or action-conditioned scene evolution, WAMs provide richer learning signals than direct imitation and offer a natural interface for incorporating visual context beyond single-frame reactive control.
A representative causal WAM, LingBot-VA~\cite{li2026causal}, retains a growing prefix of past observations and interleaves dynamics prediction with inverse-dynamics action decoding, allowing actions to exploit all accumulated visual evidence.
However, this frame-wise history incurs rapidly growing inference latency and GPU-memory costs as trajectory length increases.

\begin{figure*}[t]
\begin{center}
    \centerline{\includegraphics[width=\textwidth]{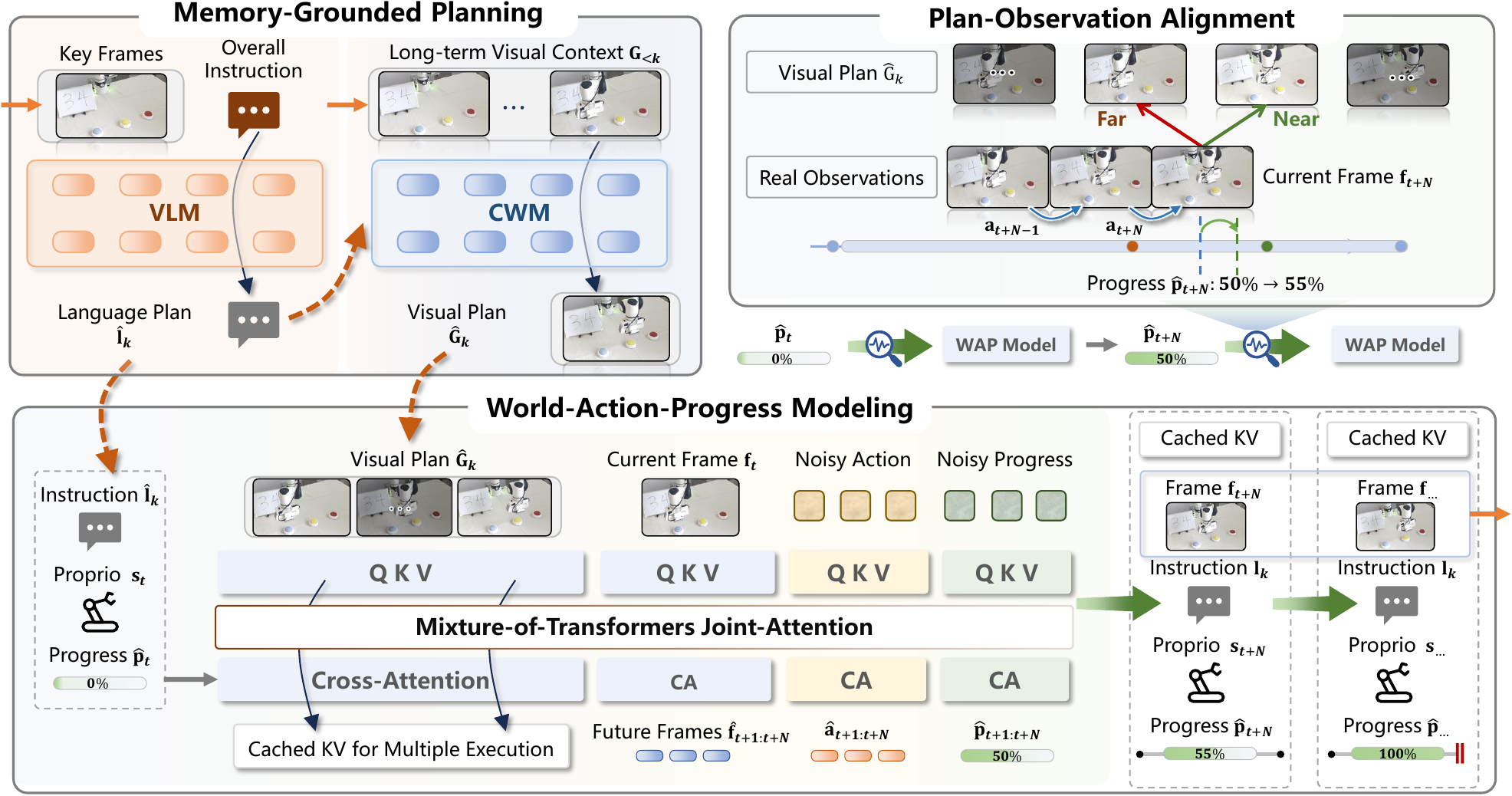}}
    \end{center}
    \vspace{-4 mm}
    \caption{\textbf{Overview of MaP-WAM.} Memory-grounded planning first predicts the next segment-level language plan $\hat{\text{l}}_k$ from the multimodal context, and then generates a visual plan $\hat{\text{G}}_k$ from the long-term visual context using a causal world model (CWM) as execution guidance.
    Conditioned on $\hat{\text{G}}_k$ and the progress condition $\hat{\text{p}}_t$, the World-Action-Progress (WAP) model jointly models future visual dynamics, actions, and task progress.
    During deployment, the fixed plan prefix is cached and reused across action chunks until the predicted progress triggers the next planning stage.
    Plan-observation alignment further retrieves visual-plan frames near the predicted progress, matches them to the current observation, and uses the best-matching plan state to calibrate progress, mitigating error accumulation from recursive prediction over long executions.
    Upon segment transition, real execution observations are resampled into sparse visual evidence and appended to the episodic context.}
    \label{fig:overview}
    \vspace{-2mm}
\end{figure*}

\noindent\textbf{Goal- and Plan-Conditioned Policies.}
Our work is more closely related to goal- and plan-conditioned policies, which predict intermediate goals or plans, represented as subgoal images~\cite{ZhaoLKFZWLMHFHL25, intelligence2026pi07}, trajectories~\cite{gu2024rttrajectory, li2025hamster}, or short videos~\cite{NEURIPS2023_1d5b9233, journals/ral/XuQS25}, and subsequently generate actions with plan-conditioned policies or inverse-dynamics models.
However, the generated goal is usually conditioned on the current observation, task instruction, or externally provided examples rather than on accumulated episodic context, and these methods typically lack a mechanism for aligning execution progress with the generated plan.
In contrast, MaP-WAM predicts memory-grounded plans and closes the loop between planning and execution through progress-aware adaptive transitions.

\vspace{-1.25mm}

\subsection{Memory Modeling for Robotic Manipulation}
Existing works on memory modeling for robotics mainly rely on language memory, continually updated memory, growing windows, or their combinations.
(1) Language Memory.
This line of work converts history into a compact language summary or an intermediate instruction before action generation.
MemER~\cite{sridhar2026scaling} and Mem-0~\cite{chen2026rmbench} select sparse visual keyframes and predict a subtask instruction for low-level VLA.
MEM~\cite{torne2026mem} maintains a recursively updated language summary as long-term memory.
Such language interfaces are compact and interpretable, but may discard fine-grained perceptual evidence.
(2) Continually Updated Memory.
These approaches maintain updatable latent states~\cite{li2024vision, li2026rememvla} or memory banks~\cite{pmlr-v267-fang25c, shi2026memoryvla, manifoldai2026worldscape} as context for action generation,
but may struggle to retain task-relevant information over long horizons, as earlier evidence can be compressed or overwritten.
(3) Growing Windows.
Other works directly extend the observation context through sliding or growing windows~\cite{guhur2023instruction, torne2026mem, chen2026rmbench, li2025cronusvla, li2026causal, DBLP:journals/corr/abs-2606-20562}.
Direct context retains richer temporal and visual evidence, but fixed windows truncate distant history, while growing windows incur increasing latency and GPU-memory costs.

\vspace{0.6mm}
\section{Our Approach}
\subsection{Overview}
\textbf{Problem Formulation.}
We formulate memory-dependent robotic manipulation as a sequential decision-making problem.
Given a language instruction $\text{l}$, the current proprioceptive state $\text{s}_t$, and the observation sequence $\text{f}_{\leq t}$, a general memory-dependent policy models
\begin{equation}
\pi(\text{a}_{t+1:t+h} \mid \text{f}_{\leq t}, \text{s}_t, \text{l}),
\end{equation}
where $\text{a}_{t+1:t+h}$ denotes the next action chunk.
The key challenge is that critical information for action generation may reside in historical observations $\text{f}_{<t}$.

\noindent\textbf{Memory-as-Plans Decomposition.}
Rather than repeatedly processing a dense, growing sequence of past observations during action generation, MaP-WAM decouples long-horizon visual-context processing from short-horizon action generation, assigning them to memory-grounded planning and plan-conditioned execution, respectively.
MaP-WAM maintains a structured multimodal episodic context $\text{C}_{<k}$ before the $k$-th segment.
This context records the execution history at the segment level, including completed segment instructions and sparsely sampled long-term visual context from previous segments.
Together with the global task instruction $\text{l}$, $\text{C}_{<k}$ provides planning-time evidence for inferring the next language plan and the desired visual evolution.
The planner models the next memory-grounded plan as
\begin{equation}
  \pi_\mathcal{P}(\text{C}_k \mid \text{C}_{<k}, \text{l}).
\end{equation}
Conditioned on the memory-grounded plan $\text{C}_k$, the execution module aims to predict short-horizon actions $\text{a}_{t+1:t+h}$ from the current observation $\text{f}_{t}$ and robot state $\text{s}_t$:
\begin{equation}
\pi_\mathcal{E}(\text{a}_{t+1:t+h}\mid \text{C}_{k}, \text{f}_{t},\text{s}_t, \text{l}), \quad [t,t+h] \subseteq \mathcal{H}(\text{C}_{k}),
\end{equation}
where $\mathcal{H}(\text{C}_{k})$ denotes the planning horizon of $\text{C}_{k}$, and the action timesteps $[t,t+h]$ lie within the temporal range covered by $\text{C}_{k}$.
This horizon is not fixed in advance but determined online by the progress-gated segment transitions described below.
$\pi_\mathcal{E}$ is rolled out repeatedly to generate actions within $\mathcal{H}(\text{C}_{k})$ until the current plan is completed.
Notably, the inputs of $\pi_\mathcal{E}$ are independent of the history length, so the executor context remains fixed as the task history grows.

\vspace{0.15mm}

\subsection{Memory-Grounded Planning}
\label{sec:planning}

\textbf{Structured Multimodal Episodic Context.}
We represent execution history as a structured multimodal episodic context of completed segment records to avoid processing a dense, ever-growing sequence of past frames.
Each completed segment $i$ contributes a record $\text{C}_i=\{\text{l}_i,\text{G}_i\}$, pairing its language instruction $\text{l}_i$ with sparse visual context $\text{G}_i$ comprising a fixed-length sequence of frames uniformly sampled from its real execution trajectory.
Additionally, the initial observation is stored as $\text{G}_0=\text{f}_0$.
The multimodal context $\text{C}_{<k}$ provides historical evidence for inferring the next language plan and desired visual evolution.

\noindent\textbf{Memory-Grounded Language-Visual Planning.}
We factorize the planning into a language planner and a visual planner.
Given the global instruction $\text{l}$, the completed segment instructions $\text{l}_{<k}$, and a compact keyframe set $\text{f}^\star$ extracted from $\text{G}_{<k}$ (the initial frame $\text{G}_0$ and the last frame of each completed segment $\text{G}_i$), the VLM planner $\pi_{\mathcal{P}}^l$ predicts the next subgoal as a language plan $\text{l}_k$:
\begin{equation}
\pi_{\mathcal{P}}^l(\text{l}_k \mid \text{l}_{<k}, \text{l}, \text{f}^\star).
\end{equation}
The resulting language plan defines the immediate semantic objective while remaining consistent with the global instruction and completed history.
We formulate $\pi_{\mathcal{P}}^v$ as a causal world model (CWM) that generates a visual plan $\text{G}_k$ as fine-grained guidance, conditioned on the long-term visual context $\text{G}_{<k}$ of completed segments, the language plan $\text{l}_k$, and the global instruction $\text{l}$.
CWM is trained with the standard flow-matching objective $\mathcal{L}_{\mathrm{FM}}$ defined in Appendix:
\begin{equation}
  \mathcal{L}_\mathcal{P}^{v} = \mathcal{L}_{\mathrm{FM}}(\text{G}_k, (\text{G}_{<k}, \text{l}_k, \text{l})),
\end{equation}
where the generation condition and target are $c=(\text{G}_{<k}, \text{l}_k, \text{l})$ and $y=\text{G}_k$, respectively.

\noindent\textbf{Causal Attention for Visual Planning.}
In the CWM, the prefix comprises a variable number of blocks corresponding to the sparse visual evidence $\text{G}_{<k}$, whereas the target block represents the future guidance $\text{G}_k$.
We organize tokens into the segment-wise blocks and apply a block-causal mask that prevents future leakage across segments and makes completed evidence a static, cacheable prefix at inference.

\subsection{World-Action-Progress Modeling}
Given the generated memory-grounded plan, the remaining challenge is to realize it over a variable and initially unknown number of control steps, owing to task complexity and stochastic execution dynamics.
To this end, we introduce the WAP model as a plan-conditioned executor and address the temporal misalignment through progress modeling, which provides an explicit alignment signal between the fixed plan and the evolving execution state, and enables adaptive planning-execution transitions.
Unlike prior work that employs progress as a post-hoc verifier or reward signal~\cite{zhang2025rewind, zhao2026tapsampling}, WAP treats progress as a first-class modality that is jointly generated with actions and fed back as a conditioning signal.

\begin{table*}[t]
\centering
\caption{\textbf{Success rates on RMBench.} Task Memory Complexity (TMC): M(1) and M(n) denote tasks requiring one and multiple task-relevant past observations, respectively. Bold and underlined entries indicate the best and second-best results.}
{\small
\setlength{\tabcolsep}{4pt}

\renewcommand{\tabularxcolumn}[1]{m{#1}}
\begin{tabularx}{\textwidth}{
@{}
>{\raggedright\arraybackslash}m{0.17\textwidth}
>{\centering\arraybackslash}m{0.05\textwidth}|
*{7}{>{\centering\arraybackslash}X}
@{}
}
\toprule
Tasks &
TMC &
DP\par\cite{DBLP:journals/ijrr/ChiXFCDBTS25} &
$\pi_{0.5}$\par\cite{black2025pi} &
X-VLA\par\cite{DBLP:journals/corr/abs-2510-10274} &
Mem-0\par\cite{chen2026rmbench} &
WLA-0\par\cite{DBLP:journals/corr/abs-2606-05979} &
LingBot-VA\par\cite{li2026causal} &
MaP-WAM\par(Ours) \\
\midrule
Observe and Pick Up & M(1) & 1\%  & 9\% & 9\% & 4\% & - & 3\% & \cellcolor[HTML]{efefff}{\textbf{19\%}} \\
Rearrange Blocks & M(1) & 0\%  & 13\% & 13\% & \cellcolor[HTML]{f9f9ff}{\underline{89\%}} & - & \cellcolor[HTML]{efefff}{\textbf{100\%}} & 66\% \\
Put Back Block & M(1) & 0\%  & 11\% & 18\% & 90\% & - & \cellcolor[HTML]{efefff}{\textbf{100\%}} & \cellcolor[HTML]{efefff}{\textbf{100\%}} \\
Swap Blocks & M(1) & 11\%  & 24\% & 16\% & 67\% & - & \cellcolor[HTML]{efefff}{\textbf{99\%}} & \cellcolor[HTML]{f9f9ff}{\underline{97\%}} \\
Swap T & M(1) & 20\%  & 15\% & 3\% & 14\% & - & \cellcolor[HTML]{f9f9ff}{\underline{88\%}} & \cellcolor[HTML]{efefff}{\textbf{96\%}} \\
\midrule
Battery Try & M(n) & 10\% & 16\% & 26\% & 28\% & \cellcolor[HTML]{f9f9ff}{\underline{45\%}} & 41\% & \cellcolor[HTML]{efefff}{\textbf{82\%}} \\
Blocks Ranking Try & M(n) & 10\% & 6\% & 1\% & 18\% & 23\% & \cellcolor[HTML]{efefff}{\textbf{100\%}} & \cellcolor[HTML]{f9f9ff}{\underline{94\%}} \\
Cover Blocks & M(n) & 0\% & 0\% & 2\% & 68\% & \cellcolor[HTML]{f9f9ff}{\underline{84\%}} & 79\% & \cellcolor[HTML]{efefff}{\textbf{100\%}} \\
Press Button & M(n) & 0\% & 0\% & 0\% & 0\% & 74\% & \cellcolor[HTML]{f9f9ff}{\underline{84\%}} & \cellcolor[HTML]{efefff}{\textbf{96\%}} \\
\midrule
\textbf{Total Average} & - & 5.8\% & 10.4\% & 9.8\% & 42.0\% & - & \cellcolor[HTML]{f9f9ff}{\underline{77.1\%}} & \cellcolor[HTML]{efefff}{\textbf{83.3\%}} \\
\bottomrule
\end{tabularx}
}
\label{tab:sim_results}
\vspace{-1.8mm}
\end{table*}

\vspace{-0.7mm}
\noindent\textbf{World-Action-Progress Model.}
We annotate each training segment $\text{f}_{i:j}$ with normalized progress $\text{p}_t=\frac{t-i}{j-i}$, where $t\in[i,j]$~\cite{zhang2025rewind, zhao2026tapsampling}.
The progress value provides a continuous coordinate for aligning execution states with the visual plan $\text{G}_k$.
As shown in Fig.~\ref{fig:overview}, we construct WAP by extending a pretrained video DiT with action and progress experts in a Mixture-of-Transformers architecture to jointly model visual dynamics, robot actions, and progress.
Given a segment plan $\text{C}_k=\{\text{l}_k,\text{G}_k\}$, the current observation $\text{f}_t$, proprioceptive state $\text{s}_t$, and current progress $\text{p}_t$, WAP jointly predicts the future visual latent, the action chunk, and the corresponding progress sequence.
WAP encodes the visual plan as a static clean prefix and the current observation as clean state tokens, while appending noisy prediction targets for visual dynamics, actions, and progress.
Its structured attention mask allows dynamic tokens to attend to the plan and current state, while keeping the plan prefix independent of dynamic tokens and cacheable throughout segment execution.
Following FastWAM~\cite{yuan2026fastwam}, we prevent action and progress tokens from attending to future visual tokens, and vice versa.
Future visual prediction therefore serves as an auxiliary world-modeling objective during training and can be omitted at inference.
In cross-attention layers, all tokens attend to the language plan $\text{l}_k$, while dynamic tokens are additionally conditioned on the proprioceptive state $\text{s}_t$ and current progress $\text{p}_t$.
The progress condition $\text{p}_t$ provides an explicit temporal anchor that disambiguates visually similar states with different semantic stages, without expanding the observation window.

\vspace{-1.5mm}
\noindent\textbf{Training Objective.}
We train WAP by applying the conditional flow-matching objective $\mathcal{L}_{\mathrm{FM}}$ to future visual states $y_v=\text{f}_{t+1:t+h}$, an action chunk $y_a=\text{a}_{t+1:t+h}$, and a progress sequence $y_p=\text{p}_{t+1:t+h}$:
\begin{equation}
  \mathcal{L}_{\mathcal{E}}^{m}
  =
  \mathcal{L}_{\mathrm{FM}}
  \left(y_m, c\right),
  \quad
  m\in\{v,a,p\},
\end{equation}
where the shared condition is $c=(\text{G}_k,\text{l}_k,\text{f}_t,\text{s}_t,\text{p}_t)$.
The final training objective is
\begin{equation}
\mathcal{L}_{\mathcal{E}} = \lambda_{v}\mathcal{L}_\mathcal{E}^{{v}} + \lambda_{a}\mathcal{L}_\mathcal{E}^{{a}} + \lambda_{p}\mathcal{L}_\mathcal{E}^{{p}},
\end{equation}
where $\lambda_{v}$, $\lambda_{a}$, and $\lambda_{p}$ are loss weights.

\subsection{Closed-Loop Planning and Execution}

\textbf{Plan-Observation Alignment for Progress Calibration.}
WAP conditions the prediction on the current progress $\text{p}_t$, while ground-truth progress is unavailable at deployment.
Therefore, the progress condition is recursively updated from WAP's predicted progress sequence, causing error accumulation over long-horizon tasks.
The visual plan, however, provides a temporally indexed visual reference, enabling progress calibration through plan-observation alignment.
As illustrated in Fig.~\ref{fig:overview}, each plan frame is indexed by normalized progress.
Given the current estimate, we retrieve nearby plan frames and select the one that is visually most similar to the observation obtained after executing the current action chunk.
The progress condition is updated toward the selected frame's progress index, anchoring execution to the planned visual evolution and improving the robustness of progress estimation.
See the Appendix for further details.

\noindent\textbf{Progress-Gated Segment Transition.}
Progress prediction provides a direct criterion that enables closed-loop planning and execution.
MaP-WAM averages the predicted progress over the latest actions and detects completion of the current segment once the resulting score exceeds a predefined threshold $\tau$, terminating execution of the current plan, updating the episodic context, and invoking memory-grounded planning for the next segment.
Specifically, the real execution observations are uniformly resampled into sparse visual context $\text{G}_k$, which replaces the generated visual plan in the appended record $\{\text{l}_k,\text{G}_k\}$, keeping the context grounded in real observations rather than generated predictions.

\section{Experiments}

\subsection{Implementation Details}

\noindent\textbf{Model Configuration.}
For language planning, we fine-tune Qwen3.5-4B \cite{qwen3.5} to predict the next segment-level language plan from the history.
For visual planning, we initialize the CWM from WAN-2.2-5B \cite{DBLP:journals/corr/abs-2503-20314} and fine-tune it using a causal input format and a block-causal attention mask.
Each completed segment is uniformly resampled into $N=8$ frames as sparse visual context.
The WAP model also uses WAN-2.2 as the video expert.
The action expert has $1.02$B parameters with hidden dimension $d_a=1024$, while the progress expert has $207$M parameters with hidden dimension $d_p=256$.

\begin{figure}
\begin{center}
\centerline{\includegraphics[width=\columnwidth]{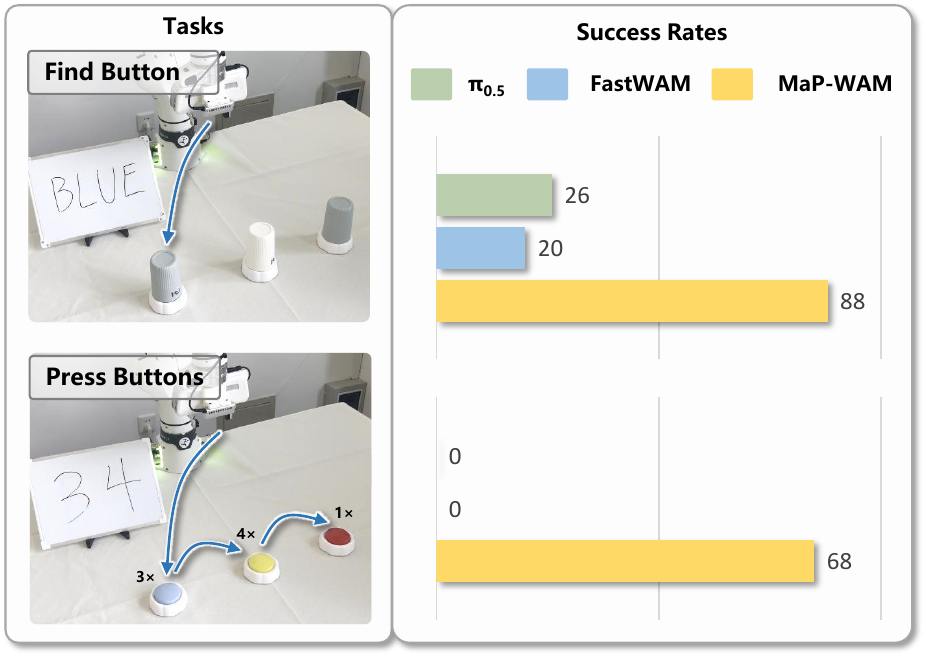}}
    \end{center}
    \vspace{-5 mm}
    \caption{\textbf{Real-world tasks and success rates.} We report the success rates over 50 trials per task.}
    \label{fig:realworld}
    \vspace{-4mm}
\end{figure}

\begin{figure*}[t]
\begin{center}
\centerline{\includegraphics[width=\textwidth]{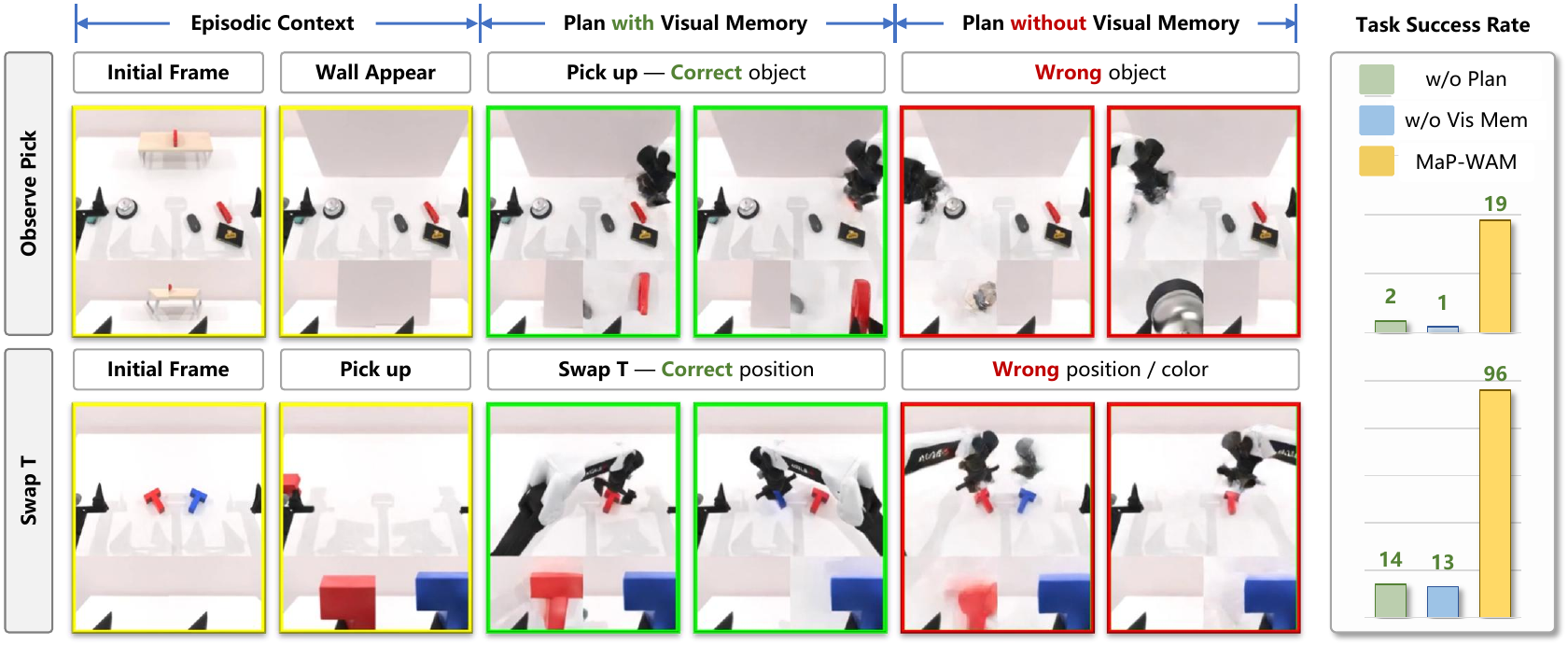}}
    \end{center}
    \vspace{-5 mm}
    \caption{\textbf{Ablation study of Memory-Grounded Visual Planning.}
    Given the same episodic context, the CWM generates visual plans with more accurate object identities and spatial configurations.
    Without visual memory, a standard current-observation-conditioned world model loses fine-grained evidence and often generates plans with wrong objects, colors, or positions.
    Visual memory is therefore particularly important for tasks requiring precise historical visual evidence.}
    \label{fig:ablation_framework}
\end{figure*}

\noindent\textbf{Training and Inference Settings.}
We set the threshold for planning-execution transition to $\tau=0.95$.
WAP is trained with the ground-truth $\text{G}_{k}$ and $\text{l}_{k}$ from the training set.
The progress condition is augmented by an additive offset sampled uniformly from $[-0.1, 0.1]$ and clipped to $[0,1]$ during training.
The loss weights are $\lambda_v=\lambda_a=\lambda_p=1.0$ for the three branches.
The action and progress branches share the same sampled flow timestep, while the future video branch uses an independent timestep.
We use 10 flow-matching denoising steps during inference for action and progress generation.
See the Appendix for further details.

\subsection{Simulation Experiments}
We evaluate MaP-WAM on RMBench \cite{chen2026rmbench}, a simulation benchmark designed for long-horizon memory-dependent robotic manipulation.
RMBench requires policies to reason over historical information that is no longer available from the current observation.
It includes five $M(1)$ tasks and four $M(n)$ tasks, corresponding to decisions that depend on one or multiple task-relevant past observations, respectively.
Following the benchmark protocol, we train MaP-WAM with $50$ official expert demonstrations per task and report success rates over $100$ evaluation rollouts per task with global seed $0$.
We compare MaP-WAM with the representative baselines evaluated on RMBench, including DP \cite{DBLP:journals/ijrr/ChiXFCDBTS25}, $\pi_{0.5}$~\cite{black2025pi}, X-VLA \cite{DBLP:journals/corr/abs-2510-10274}, Mem-0~\cite{chen2026rmbench}, WLA-0 \cite{DBLP:journals/corr/abs-2606-05979}, and LingBot-VA \cite{li2026causal}.
Our planners are trained in a multi-task setting, while WAP is trained in a single-task setting following MEM-0.

Table~\ref{tab:sim_results} shows that MaP-WAM achieves the highest overall success rate of 83.3\% across RMBench, outperforming all representative baselines.
Notably, \textit{Swap T} and \textit{Press Button} respectively require precise memory of initial object locations and disambiguation of visually similar but semantically distinct execution states.
Benefiting from memory-grounded planning and progress modeling, MaP-WAM achieves a 96\% success rate on both tasks, exceeding existing methods.
Additionally, \textit{Observe and Pick Up} is particularly challenging because it requires distinguishing among dozens of object types and jointly reasoning over temporally separated observations and spatial relations, while only 50 demonstrations are available per task.
Nevertheless, MaP-WAM improves the success rate over the strongest baseline from 9\% to 19\%.

\subsection{Real-World Experiments}
Real-world experiments are conducted on a $7$-DoF Franka Research 3 robot arm with third-person and wrist-mounted RealSense D435i cameras.
We evaluate MaP-WAM on two categories of memory-dependent manipulation tasks.
In \textit{Find Button}, the robot first observes the colors of a set of buttons. The buttons are then occluded by covers, after which a human presents a color instruction on a whiteboard.
The robot is expected to open the corresponding cover to find the button.
In \textit{Press Buttons}, the robot is expected to observe two numbers displayed on a whiteboard, press the left and middle buttons multiple times, respectively, before pressing the right confirmation button.
As shown in Fig.~\ref{fig:realworld}, we collect 50 trajectories for each task to train these methods, and MaP-WAM achieves 88\% and 68\% success rates over 50 independent trials per task.
By contrast, the baselines attain non-zero success on \textit{Find Button} through random selection, but fail on the more demanding \textit{Press Buttons} task.

\begin{table}
\centering
\caption{\textbf{Ablation study of progress modeling designs.}}
\small
\setlength{\tabcolsep}{3pt}
\begin{tabular}{@{}m{0.28\columnwidth}|*{3}{>{\centering\arraybackslash}m{0.16\columnwidth}}|>{\centering\arraybackslash}m{0.16\columnwidth}@{}}
\toprule
    & Blocks Ranking & Cover Blocks & Press Button & Average \\
\midrule
classification   & 40\%  & 71\% & 0\% & 37.0\% \\
\midrule
w/o prog. cond.  & 57\% & 87\%  & 18\% & 54.0\% \\
w/o prog. calib. & 38\% & 85\%  & \cellcolor[HTML]{efefff}{\textbf{98\%}} & 73.7\% \\
\textbf{MaP-WAM}   & \cellcolor[HTML]{efefff}{\textbf{94\%}} & \cellcolor[HTML]{efefff}{\textbf{100\%}}  & 96\% & \cellcolor[HTML]{efefff}{\textbf{96.7\%}} \\
\bottomrule
\end{tabular}
\label{tab:ablation_progress}
\vspace{-3mm}
\end{table}

\subsection{Further Analysis}

\textbf{Memory-Grounded Visual Planning.}
We evaluate the role of visual planning on two RMBench tasks, namely \textit{Observe and Pick Up} and \textit{Swap T}.
We compare MaP-WAM with two variants: \textit{w/o visual plan}, which removes the CWM planner and conditions WAP only on the VLM-derived language plan, and \textit{w/o visual memory}, which replaces the CWM with a standard current-observation-conditioned world model.
The backbone architectures of all retained components are kept identical to their counterparts in MaP-WAM for a controlled comparison.
As shown in Fig.~\ref{fig:ablation_framework},
the \textit{w/o visual plan} variant isolates the value of visual guidance, whereas \textit{w/o visual memory} tests whether such guidance must be grounded in long-horizon episodic evidence rather than the current observation.
Both variants reduce the success rate on \textit{Observe and Pick Up}, showing that its performance depends on generating visual plans from historical object evidence.
A similar pattern is observed on \textit{Swap T}, where execution requires retaining the historical locations of two T-shaped blocks before swapping them.
Replacing episodic visual evidence with the current observation leads to incorrect object identities or spatial positions, whereas memory-grounded visual planning preserves critical information.

\vspace{0.8mm}
\noindent\textbf{Progress Modeling.}
We conduct ablation experiments on progress modeling designs across three challenging RMBench tasks, namely \textit{Blocks Ranking Try}, \textit{Cover Blocks}, and \textit{Press Button}.
To assess the role of each design, we compare MaP-WAM with three variants:
(1) \textit{classification}, which removes progress modeling and instead adds tokens that predict whether the current segment is complete,
(2) \textit{w/o progress condition}, which retains progress prediction but removes the current progress $\text{p}_t$ from WAP's conditioning inputs,
and (3) \textit{w/o progress calibration}, which retains recursively predicted progress but disables plan-observation alignment at deployment.
As shown in Table~\ref{tab:ablation_progress}, the classification baseline yields the lowest average success rate across the three tasks.
Removing the progress condition substantially degrades performance on \textit{Press Button}.
This task contains visually similar pressing and releasing phases, and without the current progress condition, the executor cannot disambiguate the current state reliably, leading to missed or repeated presses.
The progress condition therefore provides an explicit temporal state signal that helps align the current observation with the intended segment execution.
Progress calibration is most beneficial on \textit{Blocks Ranking Try}, which is substantially longer than the other evaluated tasks.
In particular, the subgoal \textit{swap $a$ and $b$} requires three consecutive pick-and-place operations, resulting in an average length of $423.10$ steps, compared with $116.32$ steps for the other tasks.
The long execution horizon leads to error accumulation in autoregressive progress prediction, while progress calibration effectively mitigates this drift through plan-observation alignment.

\begin{figure}[t]
\begin{center}
    \centerline{\includegraphics[width=\columnwidth]{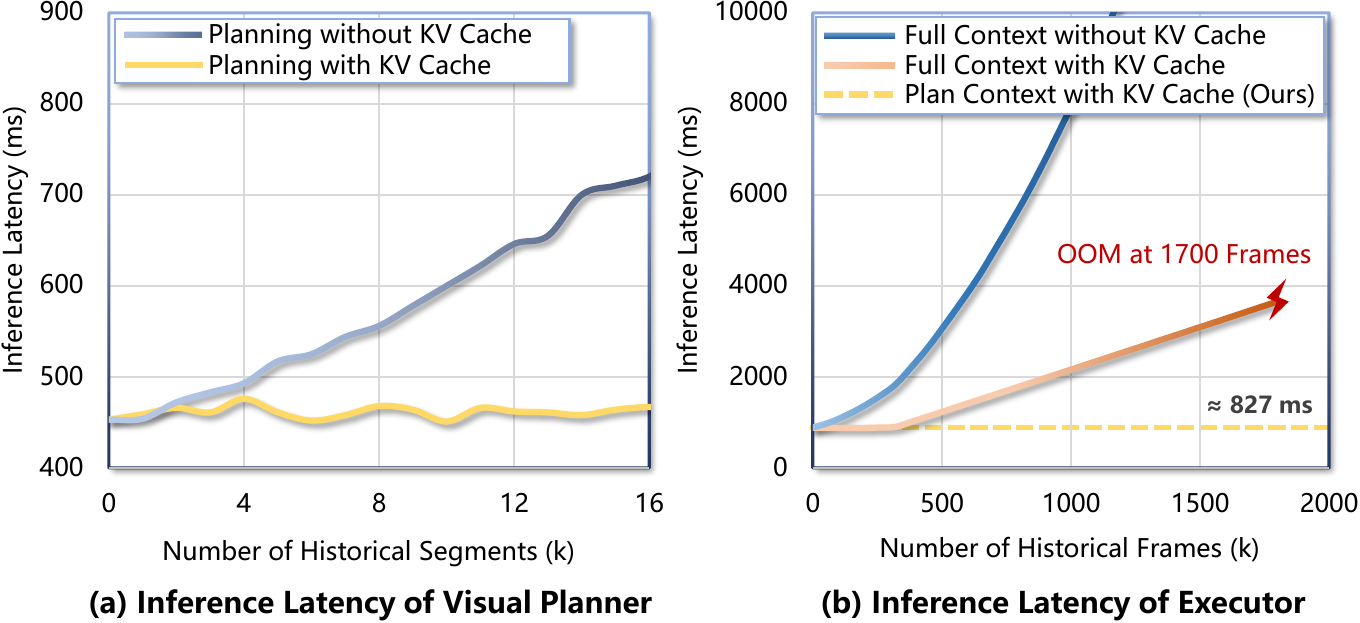}}
    \end{center}
    \vspace{-3.7 mm}
    \caption{\textbf{Inference latency as history grows.}
  (a) Visual planner inference latency versus the number of completed historical segments, with and without KV caching.
  (b) Per-chunk executor inference latency versus the number of historical frames for full-context execution and MaP-WAM's plan-context execution.}
    \label{fig:latency}
    \vspace{-4 mm}
\end{figure}

\noindent\textbf{Inference Efficiency.}
Fig.~\ref{fig:latency} reports planner inference latency as the number of completed history segments increases, and executor latency per action chunk as the number of historical frames increases.
For planning, MaP-WAM maintains a sparse visual context that retains only two visual latent timesteps per completed subtask after temporal compression by the WAN-VAE encoder \cite{DBLP:journals/corr/abs-2503-20314}.
The causal attention design of the CWM makes evidence from completed segments a cacheable episodic prefix, thereby keeping planner inference efficient even after 16 consecutive segments.
Planning is only invoked at each segment transition, after the preceding segment is completed.
In contrast, WAP is queried repeatedly within each segment to generate action chunks.
Consequently, inference latency is dominated by action generation.
For execution, we compare WAP with a \textit{Full Context} variant that conditions the same executor on the growing frame-wise observation history instead of the fixed plan prefix.
Its per-chunk latency without KV caching grows rapidly as history length increases.
KV caching reduces repeated prefix computation, but the full-context executor incurs roughly $4\times$ the zero-history latency with 1,500 historical frames.
At 1,700 history frames, the GPU memory usage exceeds 80 GB, resulting in an out-of-memory failure.
In contrast, WAP caches a fixed plan prefix, maintaining an approximately 827ms action-chunk latency across the same range.

\section{Conclusion}

We presented MaP-WAM for memory-dependent robotic manipulation, where task execution requires information beyond the current observation.
Rather than repeatedly conditioning the executor on dense visual histories, MaP-WAM converts structured episodic context into memory-grounded plans that couple task semantics with anticipated visual evolution.
By combining memory-grounded planning with progress-aware execution, MaP-WAM closes the loop among planning, execution, and context updates while preserving planning-time access to long-term, fine-grained visual evidence and enabling efficient KV-cached inference with a fixed executor context.
Experiments in simulation and on real robots show that MaP-WAM improves performance on long-horizon memory-dependent tasks while maintaining approximately constant per-chunk executor latency as task history grows.

\noindent\textbf{Limitations.}
MaP-WAM currently builds on the segment structure available in existing benchmarks to organize memory and progress.
Extending the planning to unsegmented demonstrations via automatic segment discovery is a natural next step.
Additionally, plan-observation alignment mechanism adopts a lightweight, training-free matching metric, and learned similarity measures may further improve the calibration robustness in visually complex scenes.

\bibliography{example_paper}
\bibliographystyle{icml2026}

\newpage
\appendix
\onecolumn

\title{Memory as Plans: World-Action Modeling with Memory-Grounded Planning\\(Supplementary Materials)}
\author{
}

\appendix
\section{Conditional Flow Matching}
\label{app:flow_matching}

For a target variable $y$ and condition $c$, flow matching constructs a noisy sample by interpolating $y$ with Gaussian noise $\epsilon \sim \mathcal{N}(0,I)$ at flow time $\rho \in (0,1)$:
\begin{equation}
y^\rho = (1-\rho)y + \rho\epsilon.
\end{equation}
Given a velocity-field model $v_\theta$, the standard flow-matching objective is
\begin{equation}
\mathcal{L}_{\mathrm{FM}}(y,c) = \mathbb{E}\left[\left\|v_\theta(y^\rho, c, \rho) - (\epsilon - y)\right\|_2^2\right].
\label{equ:fm}
\end{equation}

\section{Construction of Multimodal Episodic Context}
\label{app:episodic_context}

We represent each episode as a structured multimodal memory over segment-level records.
Given an observation sequence $\text{F}=\{\text{f}_0,\text{f}_{1:i-1},\text{f}_{i:j},\ldots\}$ partitioned into semantically coherent segments, with each segment $k$ associated with a single language instruction $\text{l}_k$, we summarize each completed segment using sparse visual evidence $\text{G}_k\in\mathbb{R}^{H\times W\times 3\times N}$.
To construct $\text{G}_k$, we divide the observation sequence of segment $k$ into $8$ equal temporal bins and retain the last frame in each bin, yielding an ordered sequence of $N=8$ key observations.
The initial observation is retained separately as $\text{G}_0=\text{f}_0$.
During training, the annotations provided by RMBench \cite{chen2026rmbench} determine the segment boundaries.
We define the record for each completed segment $i$ as $\text{C}_i=(\text{l}_i,\text{G}_i)$ and the episodic context available before planning segment $k$ as $\text{C}_{<k}$.
Together with the global task instruction $\text{l}$, this context provides planning-time evidence for predicting the next language plan $\text{l}_k$ and desired visual evolution $\text{G}_k$.
For language planning, the compact keyframe set comprises $\text{G}_0$ and the final frame of each completed segment, while the CWM uses the full sparse visual context $\text{G}_{<k}$.
Before segment $k$, this prefix contains $1+8(k-1)$ frames and is temporally compressed by the WAN-VAE \cite{DBLP:journals/corr/abs-2503-20314} into $1+2(k-1)$ latent timesteps.

\begin{figure}[h]
\begin{center}
    \centerline{\includegraphics[width=0.5\textwidth]{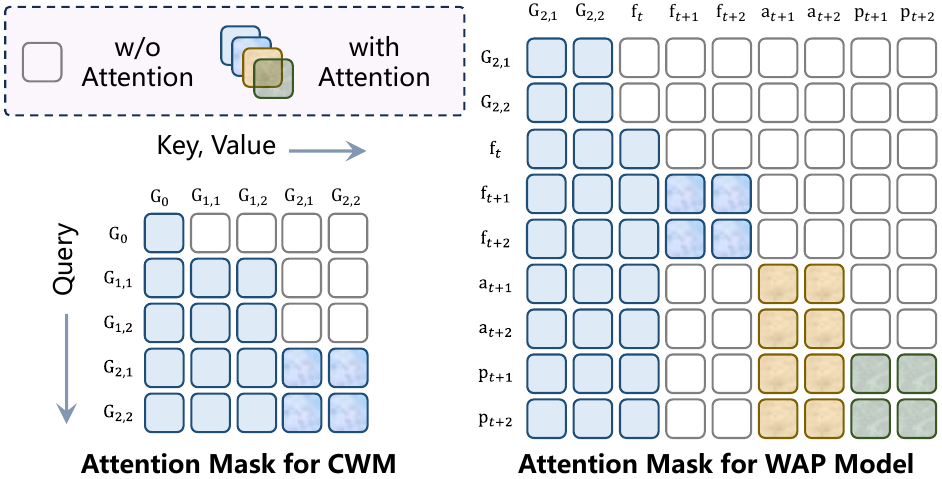}}
    \end{center}
    \vspace{-3mm}
    \caption{Self-attention masks for CWM and WAP.}    
    \vspace{-3mm}
    \label{fig:attention_mask}
\end{figure}

\section{Attention Mask and KV Caching Details}
\label{app:kv_caching}

\textbf{Causal Attention and KV Caching for CWM Planning.}
Visual evidence from completed segments $\text{G}_{<k}$ forms a causal prefix when planning segment $k$.
As shown in Fig.~\ref{fig:attention_mask}, the block causal self-attention mask allows tokens in block $i$ to attend only to blocks $j\leq i$.
In cross-attention, completed evidence blocks attend only to the global instruction $\text{l}$, whereas the target block attends to the segment language plan $\text{l}_k$.
The hidden states of the completed prefix therefore do not depend on the target block or $\text{l}_k$, allowing the corresponding key-value states to be computed once and reused at later planning stages within the same episode.

\noindent\textbf{Plan-Prefix Caching for WAP Execution.}
WAP encodes the generated visual plan $\hat{\text{G}}_k$ as a static clean prefix, followed by clean state tokens for the current observation $\text{f}_t$ and noisy target tokens for future visual latents, actions, and progress.
As shown in Fig.~\ref{fig:attention_mask}, the self-attention mask allows tokens in the plan block to attend only to the plan block, whereas tokens in the state block attend to both the plan and state blocks.
Future visual, action, and progress tokens attend to the plan and state blocks as well as tokens within their respective branches.
Progress tokens additionally attend to action tokens, while attention between future visual tokens and action or progress tokens is masked in both directions.
In cross-attention, all token groups attend to the segment-level language plan $\hat{\text{l}}_k$, whereas the state and target tokens are additionally conditioned on the proprioceptive state $\text{s}_t$ and current progress $\text{p}_t$.
This design allows WAP to prefill the plan prefix once per segment and reuse its cached key-value states throughout execution, as the plan representation is invariant to dynamic state and target tokens.
The cache is refreshed only when a progress-gated transition triggers the generation of a new segment plan.
Following FastWAM \cite{yuan2026fastwam}, we isolate the future visual branch from the action and progress branches, allowing it to be used for auxiliary training and omitted during deployment.

\begin{figure*}[t]
    \centering
    \begin{minipage}{0.9\textwidth}
        \begin{algorithm}[H]
        \DontPrintSemicolon
        \caption{MaP-WAM Inference}
        \label{alg:mapwam_inference}
        
        \KwIn{Global instruction $\text{l}$, initial observation $\text{f}_0$, and transition threshold $\tau$}
        
        Initialize $\text{G}_0\leftarrow\text{f}_0$,
        $\text{C}_{<1}\leftarrow(\text{G}_0)$,
        $k\leftarrow1$, and $t\leftarrow0$\;
        
        \While{\textnormal{the task is incomplete}}{
          Extract language planner keyframes $\text{f}^{\star}$ from $\text{C}_{<k}$\;
        
          Generate the language plan: 
          $\hat{\text{l}}_k\leftarrow
          \pi_{\mathcal{P}}^l(\text{l},\text{l}_{<k},\text{f}^{\star})$\;
        
          Generate the visual plan:
          $\hat{\text{G}}_k\leftarrow
          \pi_{\mathcal{P}}^v(\text{G}_{<k},\hat{\text{l}}_k,\text{l})$\;
        
          Form the multimodal plan:
          $\hat{\text{C}}_k\leftarrow
          (\hat{\text{l}}_k,\hat{\text{G}}_k)$\;
        
          Prefill and cache the WAP plan prefix:  $\mathcal{K}_{k}^{\mathcal E} \leftarrow \pi_{\mathcal{E}}(\hat{\text{l}}_k,\hat{\text{G}}_k)$\;
        
          Initialize: $\text{p}_t\leftarrow0$, $s\leftarrow0$, and
          $\mathcal{B}_k\leftarrow[\text{f}_t]$\;
        
          \While{$s<\tau$}{
              Calibrate the recurrent progress condition:
              $\text{p}_t\leftarrow
              \mathrm{Calibrate}(\text{f}_t,\hat{\text{G}}_k,\text{p}_t)$\;
        
              Generate action and progress sequences:
              $(\hat{\text{a}}_{t+1:t+h},\hat{\text{p}}_{t+1:t+h})
              \leftarrow
              \pi_{\mathcal{E}}(\mathcal{K}_{k}^{\mathcal E},
              \text{f}_t,\text{s}_t,\text{p}_t)$\;
        
              Execute the $h$ predicted actions and collect observations\;
              Append the collected observations to $\mathcal{B}_k$\;
        
              Compute the transition score: $s\leftarrow
              \mathrm{CompletionScore}(\hat{\text{p}}_{t+1:t+h})$\;
        
              Update state: $\text{p}_{t+h}\leftarrow\hat{\text{p}}_{t+h}$ and
              $t\leftarrow t+h$\;
          }
        
          Resample $\mathcal{B}_k$ into $N$ observations as visual evidence $\text{G}_k$\;
        
          Update episodic context: $\text{C}_{<k+1}\leftarrow
          \operatorname{Append}
          \bigl(\text{C}_{<k},(\hat{\text{l}}_k,\text{G}_k)\bigr)$\;
          $k\leftarrow k+1$\;
        }
        \end{algorithm}
    \end{minipage}
\end{figure*}

\section{Plan-Observation Alignment Details}
\label{app:progress_calibration}
At the beginning of each segment, the reference sequence comprises the initial observation of this segment and $N=8$ generated visual plan frames, with progress increasing linearly from $0$ to $1$ across the resulting $9$ frames.
Before inferring each action chunk, we calibrate the current progress estimate $\text{p}_t$ using the current observation $\text{f}_t$ and the reference sequence.
The two reference frames with progress values nearest to $\text{p}_t$ are compared with $\text{f}_t$ using mean absolute pixel difference in RGB space.
The progress condition for the upcoming WAP inference is set to the average of $\text{p}_t$ and the progress value assigned to the reference frame that is visually closer to $\text{f}_t$.
Although not necessarily optimal for visual similarity, this metric is simple and training-free.
The ablation results indicate that this lightweight calibration improves robustness to accumulated drift in recursive progress prediction over long execution horizons.

\section{Algorithm Description}
Algorithm~\ref{alg:mapwam_inference} summarizes the overall inference procedure of MaP-WAM.
At each prediction step, WAP jointly predicts an action chunk and its corresponding progress sequence.
Only a prefix of the action chunk and the corresponding prefix of the progress sequence are retained for execution, progress update, and completion score computation.
This truncation step is omitted from the algorithm for clarity.

\section{Implementation Details}
\label{app:details}
Following LingBot-VA~\cite{li2026causal}, we initialize the action and progress experts by interpolating the pretrained weights of the video expert~\cite{DBLP:journals/corr/abs-2503-20314} to match their respective hidden dimensions and rescaling the interpolated weights to preserve output variance.
For both CWM and WAP, we use $1{,}000$ flow-matching timesteps during training and set the shift coefficient of the noise-time schedule to $5.0$ during both training and inference.
At inference, both CWM and WAP use $10$ flow-matching denoising steps, with WAP simultaneously generating each action chunk and its corresponding progress sequence.
We use an action horizon of $h=32$ for WAP training and execute the first 8 actions at inference. 
All models are trained on eight NVIDIA A800 GPUs, each with $80\,\mathrm{GB}$ of memory.
The number of WAP training epochs is adjusted for each task according to the number of available training frames.
More training configurations are reported in Table~\ref{tab:training_hyperparameters}.

\begin{table}
  \centering
  \caption{Training configurations.}
  \small
  \begin{tabularx}{0.47\columnwidth}{lccc}
    \toprule
    Setting & VLM & CWM & WAP \\
    \midrule
    Per-GPU Batch Size & 4 & $1$ & $8$ \\
    Grad-Accumulation & 2 & $8$ & $8$ \\
    Optimizer & AdamW & AdamW & AdamW \\
    Learning Rate & $1\times10^{-4}$ & $1\times10^{-4}$ & $1\times10^{-4}$ \\
    LoRA Rank & 32 & - & - \\
    LoRA Alpha & 64 & - & - \\
    Training Epochs & $8$ & $100$ & $40-100$ \\
    LR Scheduler & Cosine & Cosine & Cosine \\
    Weight Decay & $0.1$ & $0.01$ & $0.01$ \\
    \bottomrule
  \end{tabularx}
  \vspace{-2mm}
  \label{tab:training_hyperparameters}
\end{table}

\section{Language-Planner Prompt Template}
\label{app:prompt}
The language planner receives an ordered image sequence comprising the initial observation and the final sampled frame from each completed segment, together with the global task instruction and the corresponding segment instructions.
The following is the prompt template.

\noindent
\fbox{%
\begin{minipage}{\columnwidth}
\small

\textbf{Question}:\\
\texttt{<image>} $\cdots$ \texttt{<image>}\\
Here is an image sequence showing the process from the robot's point of view while executing the following task: \texttt{<global task instruction>}.\\
Image $1$ shows the initial environment. The completed subtask sequence is:\\
$1$. \texttt{<instruction of segment 1>}\\
$2$. \texttt{<instruction of segment 2>}\\
$\vdots$\\
$k-1$. \texttt{<instruction of segment k-1>}\\
The results after these subtasks are shown in Image $2$, Image $3$, $\ldots$, and Image $k$, respectively. Analyze what subtask $k$ is.
\par\vspace{0.2pt}
\noindent\rule{\linewidth}{0.4pt}
\par\vspace{1pt}
\textbf{Answer:}\\
{subtask $k$: \texttt{<instruction of segment $k$>}}
\end{minipage}%
}

\section{Latency Measurement Protocol}
\label{app:latency_protocol}

All inference latency measurements are conducted on a single NVIDIA A800-SXM4 GPU with $80\,\mathrm{GB}$ of memory using PyTorch 2.7.1 and CUDA 11.8, with BF16 precision, a batch size of one, and $10$ flow-matching denoising steps.
The reported measurements capture latent-space model inference and exclude model loading, prompt encoding, data transfer, and VAE encoding or decoding.
For each configuration, we perform five warm-up runs followed by $100$ measured runs.
CUDA is synchronized immediately before and after each model invocation, and the elapsed wall-clock time is averaged over the $100$ measured runs.
For cached inference, the prefix is prefilled before timing, so the reported latency reflects steady-state inference with an established cache.
All compared configurations use the same model backbone.

\end{document}